\documentclass[10pt,twocolumn,letterpaper]{article}

\usepackage{cvpr}
\usepackage{times}
\usepackage{epsfig}
\usepackage{graphicx}
\usepackage{amsmath}
\usepackage{amssymb}
\usepackage{physics}
\usepackage{algorithm}
\usepackage{algorithmic}
\usepackage{bm}
\usepackage{booktabs,colortbl,multirow}
\usepackage{subfig}
\usepackage[numbers,sort&compress]{natbib}
\usepackage{afterpage}
\usepackage{dblfloatfix}
\usepackage{xcolor}
\usepackage{wrapfig}
\usepackage{url}

\def\etal{\emph{et al.~}}

\cvprfinalcopy 

\ifcvprfinal\fi
\begin{document}

\title{Re-Identification with Consistent Attentive Siamese Networks}

\author{Meng Zheng$^{1}$, Srikrishna Karanam$^{2}$, Ziyan Wu$^{2}$, and  Richard J. Radke$^{1}$\\
$^{1}$Department of Electrical, Computer, and Systems Engineering, Rensselaer Polytechnic Institute, Troy NY\\
$^{2}$Siemens Corporate Technology, Princeton NJ\\
{\tt\small zhengm3@rpi.edu,\{first.last\}@siemens.com,rjradke@ecse.rpi.edu}
}

\maketitle
\thispagestyle{empty}

\begin{abstract}
   We propose a new deep architecture for person re-identification (re-id). While re-id has seen much recent progress, spatial localization and view-invariant representation learning for robust cross-view matching remain key, unsolved problems. We address these questions by means of a new attention-driven Siamese learning architecture, called the Consistent Attentive Siamese Network. Our key innovations compared to existing, competing methods include (a) a flexible framework design that produces attention with only identity labels as supervision, (b) explicit mechanisms to enforce attention consistency among images of the same person, and (c) a new Siamese framework that integrates attention and attention consistency, producing principled supervisory signals as well as the first mechanism that can explain the reasoning behind the Siamese framework's predictions. We conduct extensive evaluations on the CUHK03-NP, DukeMTMC-ReID, and Market-1501 datasets and report competitive performance. 
\end{abstract}

\section{Introduction}

Given an image or a set of images of a person of interest in a ``probe" camera view, person re-identification (re-id) attempts to retrieve this person of interest among a set of ``gallery" candidates in another camera view. Due to its broad appeal in several video analytics applications such as surveillance, re-id has seen explosive growth in the computer vision community \cite{KaranamBenchmark_PAMI17, Zheng_overview_CoRR16, Market1501_ICCV15}.

While we have seen tremendous progress in re-id \cite{Dapeng_ECCV18, Yumin_ECCV18, sunPCB_ECCV18, MLFN_CVPR18, DaRe_CVPR18, DuATM_CVPR18, SuPose_CVPR17,  MGN_MM18}, there are several problems that still hinder the reliable, real-world use of person re-id. Probe and gallery camera views in real-world applications typically have large viewpoint variations, causing substantial view misalignment between probe and gallery images of the same person. Illumination differences between the locations where the cameras are installed, as well as occlusions in the captured data, add to re-id's challenges. Ideally, we want a method that can reliably spatially localize the person of interest in the image, while also providing a robust representation of the localized part in order to match accurately to the gallery of candidates. This suggests we consider the spatial localization and feature representation problems jointly and formulate the learning objective in a way that can facilitate end-to-end learning. 

\begin{figure}[t!]
	\centering
	\includegraphics[width=0.95\linewidth]{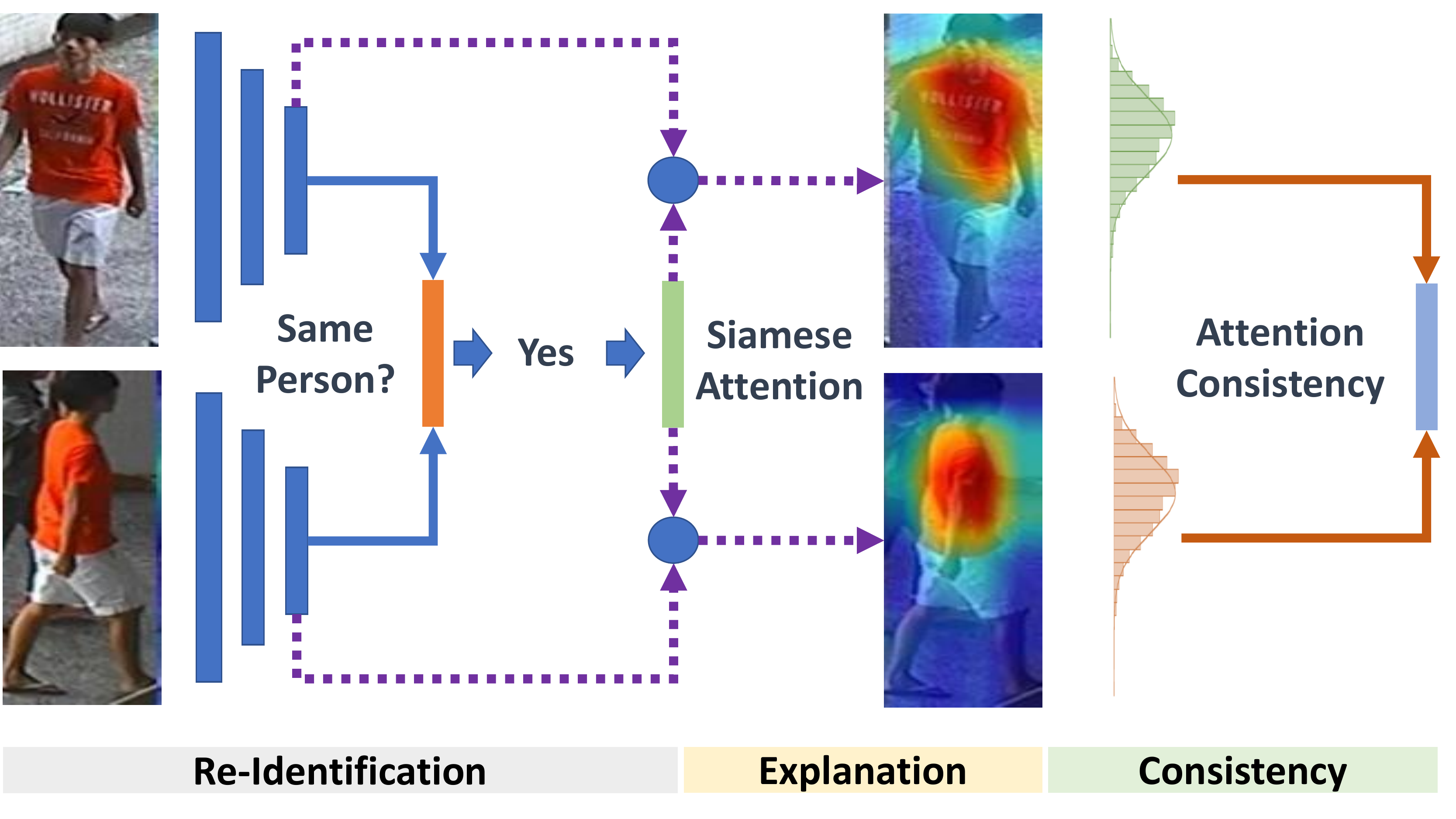}%
	\vspace{-0.25cm}
	\caption{We present the first framework for re-id that provides mechanisms to make attention and attention consistency end-to-end trainable in a Siamese learning architecture, resulting in a technique for robust cross-view matching as well as explaining the reasoning for why the model predicts that the two images belong to the same person.} 
	\vspace{-0.65cm}
	\label{fig:teaser}
\end{figure}

Attention is a powerful concept for understanding and interpreting neural network decisions \cite{weakly_CVPR15, CAM_CVPR16, Weakly_PAMI17, SelGradCAM_ICCV17}, providing ways to generate attentive regions given image-level labels and trained models, and to perform spatial localization. Unlike its use as a weight matrix in some existing work \cite{bahdanau2014neural,andreas2016neural,vaswani2017attention}, here we refer to attention computed by means of class-specific gradient backpropagation \cite{CAM_CVPR16,SelGradCAM_ICCV17}. Some recent extensions \cite{GAIN_CVPR18} take this a step forward  by training models with attention providing end-to-end supervision, resulting in improved spatial localization. These methods were not designed for the re-id problem and consequently did not have to consider localization and invariant representation learning jointly. While there have been some attempts at joint learning with these two objectives \cite{wu2018and, LiMSCAN_CVPR17, liuHPnet_ICCV17, LiHACNN_CVPR18}, these methods do not explicitly enforce any sort of attention consistency  between images of the same person. Intuitively, given same-person images from different views, there typically exist some common regions that are important for matching, which should be reflected in how attention is modeled and used for supervision. 

Furthermore, such attention consistency should lead to consistent feature representations for the two different images, leading to invariant representations for robust cross-view matching. These considerations naturally suggest the design of a Siamese framework that jointly learns consistent attention regions for images of the same person while also producing robust, invariant feature representations. While one recent paper approached these problems jointly \cite{wu2018and}, this method requires specially-designed architectures for attention modeling and considers the attention in each image independently, ignoring the intuition that attentive regions across images of the same person have to be consistent.  It also does not have an explicit mechanism to explain the reasoning behind the model's prediction. To this end, we design and propose a new deep architecture for re-id, which we call the Consistent Attentive Siamese Network (CASN), addressing all the key questions and considerations discussed above (Figure \ref{fig:teaser}). Specifically, we design a novel two-branch architecture that (a) produces attentive regions during training without requiring any additional supervision other than identity labels or any specially-designed architecture for modeling attention, (b) explicitly enforces these attentive regions to be consistent for the same person, (c) uses attention and attention consistency as an explicit and principled part of the learning process, and (d) learns to produce robust representations for cross-view matching.

To summarize, our key contributions include:

\begin{itemize}
    \item We present a technique that makes spatial localization of the person of interest a principled part of the learning process, providing supervision only by means of  person identity labels.  This makes spatial localization end-to-end trainable and automatically discovers complete attentive regions. 
    \item We present a new scheme that enforces attention consistency as part of the learning process, providing supervision that facilitates end-to-end learning of consistent attentive regions of images of the same person.
    \item We present the first learning architecture that integrates attention consistency and Siamese learning in a joint learning framework.
    \item We present the first Siamese attention mechanism that jointly models consistent attention across similar images, resulting in a powerful method that can help explain the reasoning behind the network's prediction.
\end{itemize}

\section{Related Work}
Traditional person re-id algorithms involved hand-crafted feature design followed by supervised distance metric learning. See Karanam \etal \cite{KaranamBenchmark_PAMI17} and Zheng \etal \cite{Zheng_overview_CoRR16} for detailed experimental and algorithmic studies. 

Recent developments in deep learning \cite{Resnet_CVPR16, huangDense_ICCV17} have influenced the design of re-id algorithms as well, with deep re-id algorithms achieving impressive performance on challenging datasets \cite{Yumin_ECCV18, sunPCB_ECCV18, Dapeng_ECCV18}. However, naive training of re-id models without being spatial-localization-aware will not result in satisfactory performance due to cross-view misalignment, occlusions, and clutter. To get around these issues, several recent methods adopt some form of localized representation learning. Zhao \etal \cite{ZhaoDLPartAlign_ICCV17} decomposed person images into different part regions and learned region-specific representations followed by an aggregation scheme to produce the overall image representation. Li \etal \cite{LiMSCAN_CVPR17} proposed to first learn and localize part body features by means of spatial transformer networks \cite{STN_NIPS15}, followed by a combination of local and global features to learn a classification network. Su \etal \cite{SuPose_CVPR17} used human pose information as a supervisory signal to learn normalized human part representations as part of an identification network. However, these and several other recent methods \cite{GAIN_CVPR18} consider the spatial localization problem in itself and produce representations and localizations that are not cross-view consistent. On the other hand, our approach tackles spatial localization and representation learning in a holistic, joint framework while enforcing consistency, which is key to re-id.

Attention has been used in re-id to tackle localization and misalignment problems. Liu \etal \cite{liuHPnet_ICCV17} proposed the HydraPlus-Net architecture that learns to discover low- and semantic-level attentive features for richer image representations. Li \etal \cite{LiHACNN_CVPR18} designed a scheme to simultaneously learn ``hard" region-level and ``soft" pixel-level attentive features for a multi-granular feature representation. Li \etal \cite{LiDiversity_CVPR18} learned multiple, predefined attention models and showed that each model corresponds to a specific body part, the outputs of which are then aggregated by means of a temporal attention model. These methods typically have inflexible region-specific attention models as part of the overall framework to learn important regions in the image, and more importantly, do not have an explicit mechanism to enforce attention consistency. Our approach is markedly different from these and other methods \cite{xu2018attention,song2018mask} in this category in that we only need image-level labels to learn attention, while also enforcing attention consistency by making it a principled part of the learning process. 

Consistency is an important aspect of re-id to account for cross-view differences. While this has been studied under the term ``equivariance" in some prior work \cite{lenc2016learning}, for re-id, it has been reflected in Siamese-like designs that attempt to learn invariant feature representations \cite{deepReID_CVPR14,triplet_CVPR16,ChenQuad_CVPR17,DuATM_CVPR18, SGGNN_ECCV18}. These models learn features and distance metrics jointly and do not address the spatial localization problem directly, typically formulating a local parts-based approach to solve the problem. In scenarios involving occlusion and clutter, this may not be an optimal solution, with attention leading to better spatial localization. To this end, our method, as opposed to these approaches, exploits attention during the learning process while also learning consistent spatial localization and invariant feature representations jointly.

\section{The Consistent Attentive Siamese Network}

\begin{figure}[hbtp!]
	\centering
		\includegraphics[width=\linewidth]{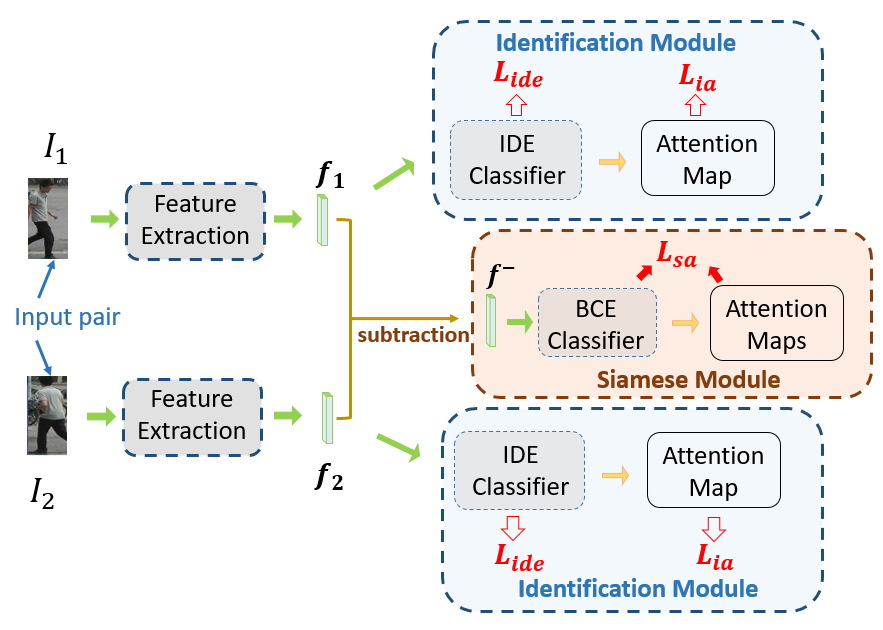}%
	\caption{The Consistent Attentive Siamese Network.}
	\vspace{-0.7cm}
	\label{fig:intro}
\end{figure}

In this section, we introduce our proposed attention-based deep architecture for person re-id, the Consistent Attentive Siamese Network (CASN), summarized in Figure \ref{fig:intro}. CASN includes an identification module and a Siamese module that provide for a powerful, flexible approach to deal with viewpoint variations, occlusions, and background clutter. The identification module (Section \ref{sec:identi_module}), with its explicit attention guidance as supervision given only identity labels, helps find reliable and accurate spatial localization for the person of interest in the image and performs identity (ID) prediction. The Siamese module (Section \ref{sec:Siamese_module}) provides the network with supervisory signals from attention consistency, ensuring that we obtain spatially consistent attention regions for images of the same person, as well as learning view-invariant feature representations for robust gallery matching. In the following, we describe each of these two modules in more detail, leading up to the overall design of the CASN.

\subsection{The Identification Module}
\label{sec:identi_module}

We first introduce the architecture of the identification module of the CASN. We begin by describing the baseline architecture for training an identification (IDE) model \cite{Zheng_overview_CoRR16}, followed by the overall identification module that integrates attention guidance into the IDE architecture. 

\subsubsection{The IDE Baseline Architecture}
\label{sec:ide_bas}

\begin{figure}[hbtp!]
	\centering
		\includegraphics[width=\linewidth]{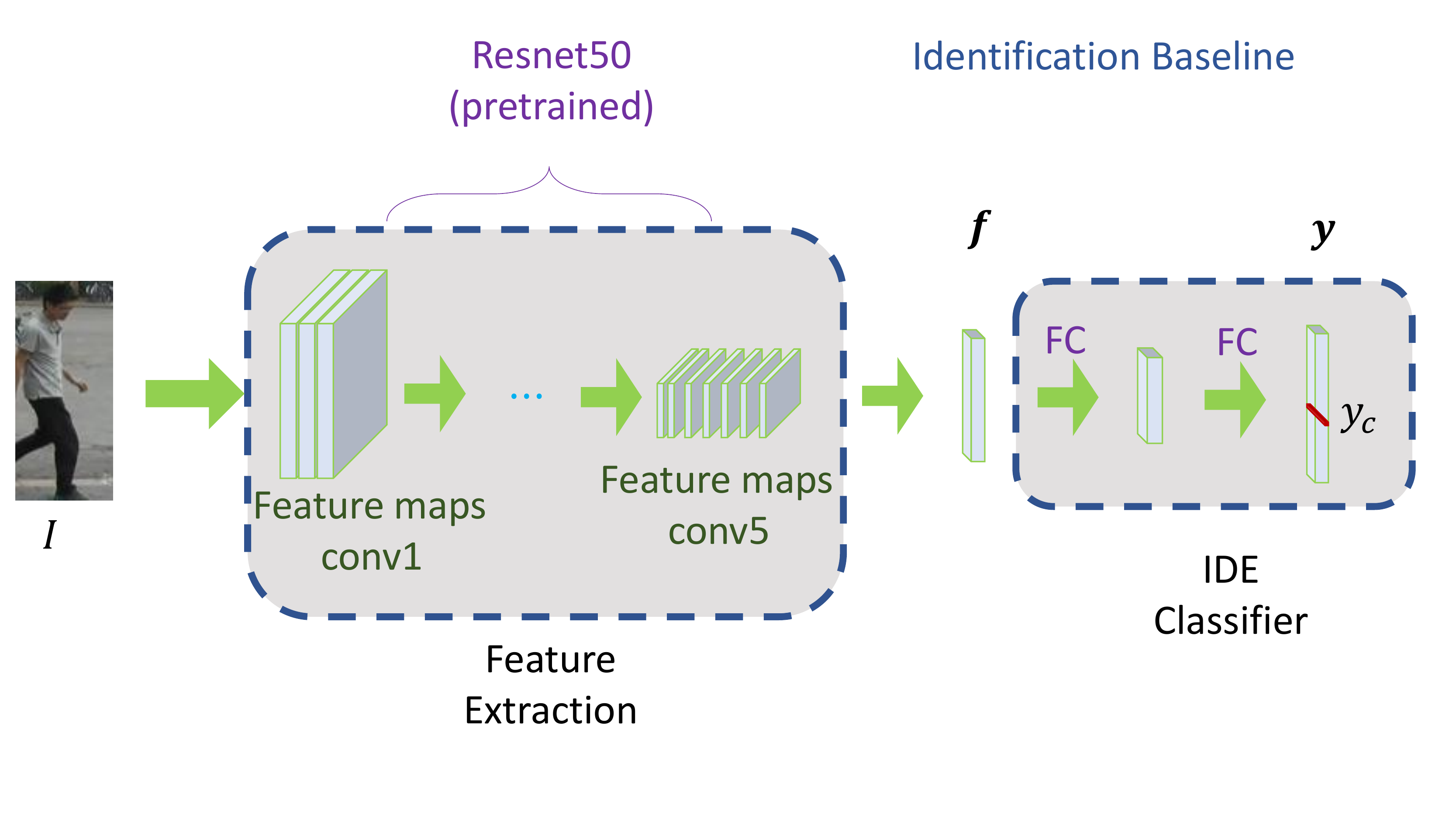}%
		\vspace{-0.7cm}
	\caption{The baseline. $\bm{f}$ is the feature vector after Resnet50 \textit{conv5}, $\bm{y}$ is the ID prediction vector with dimensionality equal to the total number of training identities, and $y_c$ is the prediction score of ID label $c$ for the input image. Note that the ``Feature extraction" block here can come from any baseline architecture, e.g., IDE or PCB \cite{sunPCB_ECCV18}.} 
	\vspace{-0.5cm}
	\label{fig:baseline}
\end{figure}

The IDE baseline is based on the ResNet50 architecture \cite{Resnet_CVPR16}, following the work in \cite{Zheng_overview_CoRR16} and recent papers that adopt ResNet50 \cite{LiDiversity_CVPR18, sunPCB_ECCV18, MGN_MM18}. Convolutional layers from \textit{conv1} through \textit{conv5} are pretrained on ImageNet \cite{imagenet_cvpr09}, following which an IDE classifier comprised of two fully-connected layers produces the identity prediction for the input image. The identification baseline is visually summarized in Figure \ref{fig:baseline}. Note that while Figure \ref{fig:baseline} shows the IDE architecture \cite{Zheng_overview_CoRR16}, this can be easily swapped with any other baseline architecture that can give the feature vector $\bm{f}$. For instance, to use the part-based convolutional baseline (PCB) architecture \cite{sunPCB_ECCV18}, one would simply swap the ``Feature Extraction" block in Figure \ref{fig:baseline} with PCB's backbone prior to obtaining $\bm{f}$. PCB is a modification of IDE that replaces the global average pooling operation in IDE with spatial pooling for discriminative part-informed feature learning. The baseline model is learned by optimizing the identification loss, which essentially maximizes the likelihood of predicting the correct class (identity) label for each training image. Formally, given $N$ training images $\{I_n\}_{n=1}^{N}$ belonging to $C$ different identities, with each image having an identity label $\{c_n\}_{n=1}^{N} \in \{1,...,C\}$, we optimize the following multi-class cross-entropy loss:
\begin{equation}
	L_{ide} = - \sum_{n=1}^N \log \frac{\exp(y_{c_n})}{\sum_j \exp(y_j)}
\label{eq:loss_ide}
\end{equation}
where $y_{c_n}$ is the prediction of class $c_n$ from the IDE classifier for input image $I_n$.

\subsubsection{Identification Attention}
\label{sec:AttMec}
Spatial localization of the person of interest is a key first step for a re-id algorithm, which should be reflected in the end-to-end learning process. While much recent work has focused on generating attention regions given image-level labels \cite{weakly_CVPR15, CAM_CVPR16, Weakly_PAMI17, SelGradCAM_ICCV17}, we need to make attention an explicit part of the learning process itself, which can then guide the network to better localize the person of interest. 

To this end, we adopt the framework of Li \etal \cite{GAIN_CVPR18} and introduce attention learning as part of our identification module, helping the network generate spatially attentive regions in person images without needing any extra information as supervision other than identity labels, which are already available.

\begin{wrapfigure}{L}{0.2\textwidth}
\centering
\includegraphics[width=0.22\textwidth]{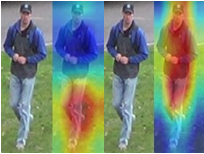}
\caption{\label{fig:IDEAtt_Demo1} An attention map with identification loss (left) and identification loss with attention learning (right).}
\end{wrapfigure}
\paragraph{}
\vspace*{-\parskip}Given an input image $I_n$ with its identity label $c_n$, we first obtain the attention (localization) map from the IDE classifier prediction by means of Grad-CAM \cite{SelGradCAM_ICCV17}. However, a re-id model trained only with IDE loss would focus only on the most discriminative regions important for satisfying the current classification objective, and may not generalize well. To better illustrate this concept, consider the Grad-CAM attention map example shown in Figure \ref{fig:IDEAtt_Demo1} (left) for an image from Market1501 \cite{Market1501_ICCV15}. The gray pants of the person attract the most attention, but the blue jacket is also useful information that is ignored in the attention map on the left. To obtain more complete attention maps and focus on the foreground subject, we use the notion of attention learning. Specifically, given $I_n$ and $c_n$, we compute its attention map $M_n$ and mask out the most discriminative regions in $I_n$ (corresponding to high responses in $M_n$) by means of the soft-masking operation $\Sigma(\cdotp)$ to get $\overline{I_n} = I_n * (1 - \Sigma(M_n))$, where $*$ is pixel-wise multiplication and $\Sigma(\cdotp) = \text{sigmoid} (\alpha(M_n - \beta))$. This produces an $\overline{I_n}$ that excludes all high-response image pixels. If $M_n$ perfectly spatially localizes the person of interest, $\overline{I_n}$ will contain no pixels contributing to the corresponding identity prediction $\overline{y_{c_n}}$. We use this notion to provide supervision to the identification module to produce more complete spatial localization. Specifically, we define the identification attention loss $L_{ia}$ for the identification module as the prediction score of masked input image $\overline{I_n}$:
\begin{equation}
    L_{ia} = \overline{y_{c_n}}
\label{eq:Mining}
\end{equation}

A comparison of the attention maps retrieved from a model trained only with the identification loss and one with identification loss and attention learning is shown in Figure \ref{fig:IDEAtt_Demo1}, where we see more foreground subject coverage with attention learning on the right. To summarize, in the identification module, we first use the IDE baseline architecture to obtain identity predictions. Attention maps are computed with Grad-CAM and refined using the identification attention objective on masked images that exclude high-attention regions to perform more complete spatial localization.

\subsubsection{Discussion}
\label{sec:IDE_discuss}
While the IDE architecture can provide a good baseline feature representation for matching \cite{Zheng_overview_CoRR16, KaranamBenchmark_PAMI17, MGN_MM18} and our proposed identification module discussed above can further lead to reasonable spatial localization by design, several problems still remain unaddressed. First, the identification module has no mechanism to ensure we obtain consistent attention regions for different images of the same person. This can be inferred from the design itself, which lacks any guiding principle to result in attention consistency. Intuitively, this is key to robust re-id since there are typically common regions in different images of the same person that need to be brought out as important during matching. Second, the identification module has no mechanism to learn invariant identity-aware representations across different camera views. Furthermore, attention consistency should correspond to consistent feature representations, suggesting it should inform representation learning. Finally, the attention component of the identification module is not particularly suitable during inference since we do not know the identity of a test image to compute its attention map. While a workaround to this problem would be to use the top-k predictions to compute attention, this clearly would be a sub-optimal solution.

The problems with the identification module lead us to the design of the Siamese module of the CASN, which attempts to address these issues in a principled manner.

\subsection{The Siamese Module}
\label{sec:Siamese_module}
In this section, we introduce the Siamese module to complement the identification module of the proposed CASN. Given a pair of input images, we first consider a binary classification problem (Section \ref{sec:BCE}), whose objective function is then used to formulate a Siamese attention mechanism (Section \ref{sec:SiaAtt}) to enforce attention consistency and consistency-aware invariant representation learning. 

\subsubsection{Binary Classification}
\label{sec:BCE}
Given a pair of input images, we construct a binary classification objective for predicting whether or not the pair belongs to the same class. Given feature vectors  $\bm{f_1}$ and $\bm{f_2}$ for the images $I_1$ and $I_2$ in the input pair (see Figure \ref{fig:baseline}), we compute the difference $\bm{f^-} = \bm{f_1} -\bm{f_2}$, which forms the input for a classifier that uses the binary cross-entropy objective (BCE) to get the class prediction for the current input pair. Note that since we set out to compute attention in the spirit of GradCAM \cite{SelGradCAM_ICCV17}, we needed a classification objective to compute Siamese attention described next, and we chose BCE for this purpose. The BCE classifier is structurally similar to the IDE classifier in Section \ref{sec:ide_bas}, with two fully connected layers. The output prediction vector $\bm{z}$ of the BCE classifier is a 2-dimensional vector, which indicates whether or not the input pair belongs to the same identity. The BCE classification objective that is optimized is defined, for a batch of $P$ input pairs, as:
\begin{equation}
\begin{split}
    &L_{bce} = -\sum_p \log \left(\frac{\exp(z_{c_p})}{\exp(z_{0}) + \exp(z_{1})}\right)\\ &~~c_p\in \{0,1\}, ~p=1,\ldots,P\\
\end{split}
\label{eq:BCEloss}
\end{equation}
where $z_{c_p}$ is the same ($c_p=1$) or different ($c_p=0$) identity prediction of the BCE classifier for input pair $p$.

\subsubsection{The Siamese Attention Mechanism} 
\label{sec:SiaAtt}

As discussed previously, identification attention alone does not ensure attention consistency and identity-aware invariant representations. To this end, we propose a new Siamese attention mechanism with explicit guidance towards attention consistency.
Consider two images $I_1$ and $I_2$ of the same identity and the corresponding BCE classifier prediction $z_1$. We first localize the attentive regions in the two images that contribute to this BCE prediction. To this end, we compute the gradient of the prediction score with respect to the feature vector $\bm{f^-}$, i.e., $\frac{\partial z_1}{\partial \bm{f^-}}$. We then find the features in $\bm{f^-}$ that have a positive influence on the final BCE prediction by means of an indicator vector $\bm{\alpha}$ constructed as:

\begin{equation}
    \alpha_i= \begin{cases}
      1, & \text{if}\ \frac{\partial z_1}{\partial f_i^-} > 0 \\
      0, & \text{otherwise}
     \end{cases}, ~i = \{0,...,\text{dim}(\bm{f^-})\}
\label{eq:alpha}
\end{equation}

Based on the indicator vector $\bm{\alpha}$, the importance scores for the input feature vectors $\bm{f_1}$ and $\bm{f_2}$ can be calculated as the dot products of $\bm{\alpha}$ and the feature vectors: $s_{1} = (\bm{\alpha}, \bm{f_1})$ and $s_{2} = (\bm{\alpha}, \bm{f_2})$. In the same spirit as Grad-CAM \cite{SelGradCAM_ICCV17}, gradients backpropagated from $s_1$ and $s_2$ are first globally average-pooled to find the channel importance weights $\alpha^k_1=\text{GAP}\left(\frac{\partial s_1}{\partial A_1}\right)$ and $\alpha^k_2=\text{GAP}\left(\frac{\partial s_2}{\partial A_2}\right)$, where $A_1$ and $A_2$ are feature maps of image $I_1$ and $I_2$ at the last convolutional layer. The attention maps can then be computed as $M_{1} = \text{ReLU}\left(\sum_k \alpha^k_1A_1^k\right)$ and $M_{2} = \text{ReLU}\left(\sum_k \alpha^k_2A_2^k\right)$.

\begin{figure}[thbp!]
	\centering
	\subfloat[]{\includegraphics[width=0.3\linewidth, height=1.5cm]{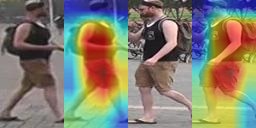}}%
	\quad
	\subfloat[]{\includegraphics[width=0.3\linewidth, height=1.5cm]{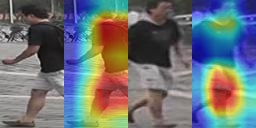}}%
	\quad
    \subfloat[]{\includegraphics[width=0.3\linewidth, height=1.5cm]{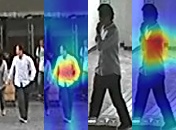}}\\%
    \subfloat[]{\includegraphics[width=0.3\linewidth, height=1.5cm]{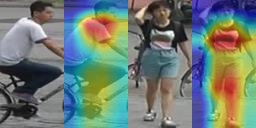}}%
	\quad
	\subfloat[]{\includegraphics[width=0.3\linewidth, height=1.5cm]{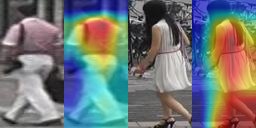}}%
	\quad
    \subfloat[]{\includegraphics[width=0.3\linewidth, height=1.5cm]{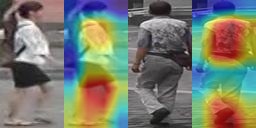}}\\%
	\caption{Demonstration of attention maps from BCE loss. (a-c): positive pairs, (d-f): negative pairs.}
	\label{fig:BCEAtt_Demo2}
\end{figure}

Visualizations of the attention maps, extracted from the BCE loss, are shown in Figure \ref{fig:BCEAtt_Demo2}. For images of the same person, we want the attention maps $M_1$ and $M_2$ to provide consistent importance to corresponding regions in the images. For instance, as we can see in Figure \ref{fig:BCEAtt_Demo2}(b), the attention map in Image 1 focuses on the full body of the person while the one in Image 2 mostly focuses on the lower part. To provide an explicit attention-consistency-aware supervisory signal and guide the network to discover consistent cross-view importance regions, we introduce the notion of spatial attention constraints based on the attention maps derived from the BCE classification objective. 

Given the attention maps $M_1$ and $M_2$, we first apply the max-pooling operation to compute the highest response across each horizontal row of pixels, giving us the two importance vectors $M_{m1}$ and $M_{m2}$. To enforce attention consistency, we explicitly constrain them to be as close as possible. To avoid alignment issues as in Figure \ref{fig:BCEAtt_Demo2}(c), we find the first and the last element of the vertical vector larger than a certain threshold $t$ in $M_{m1}$ and $M_{m2}$, and then resize the remaining elements to be of the same dimensions. We define the Siamese attention loss that enforces attention consistency as:
\begin{equation}
\begin{split}
    &L_{sa} = L_{bce} + \alpha \|M^*_{m1}-M^*_{m2}\|_2 \\
\end{split}
\label{eq:Spatial_cons}
\end{equation}
where $L_{bce}$ is defined in Equation \ref{eq:BCEloss}, $M^*_{m1}$ and $M^*_{m2}$ are resized vectors of $M_{m1}$ and $M_{m2}$ after alignment, $\|M^*_{m1}-M^*_{m2}\|_2$ is the $l_2$ distance between $M^*_{m1}$ and $M^*_{m2}$, and $\alpha$ is a weight parameter controlling the importance of the BCE loss vis-a-vis the spatial attention constraints.

\begin{figure*}[ht]
	\centering
		\includegraphics[width=0.9\linewidth]{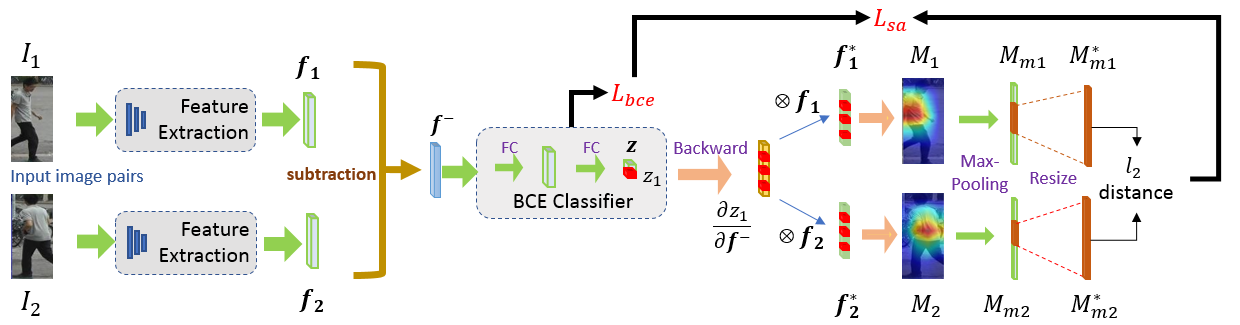}%
	\caption{Demonstration of the Siamese Attention Mechanism. Yellow arrows denote backward operation and green arrows denote forward operation. The BCE loss $L_{bce}$ and spatial constraints are added as Siamese Attention loss $L_{sa}$. Note that the ``Feature extraction" block here can come from any baseline architecture, e.g., IDE or PCB.}
	\vspace*{-0.5cm}
	\label{fig:BCE_Att}
\end{figure*}

A visual summary of our proposed Siamese attention mechanism is shown in Figure \ref{fig:BCE_Att}. For input pairs belonging to the same identity, attention maps are retrieved from the BCE classifier predictions, following which they are max-pooled to gather localization statistics for enforcing spatial attention consistency. 

\subsection{Overall Design of the CASN}
With the identification and Siamese modules discussed in the previous sections, we now present our overall framework that integrates these two modules. Our proposed CASN, depicted in Figure \ref{fig:intro}, is a two-branch architecture. During training, we pass as input a pair of images belonging either to the same or different identity. After feature extraction (see Figure \ref{fig:baseline}), the feature vectors are input to the identification module and Siamese module separately. In the identification module, the feature vectors are first passed to the IDE classifier for identity classification, following which an attention map for the input image in the current branch is retrieved from its identity label. The identification attention loss then guides the identification module to discover complete attention regions for the input image. The Siamese module takes as input the element-wise subtraction of the feature vectors from two branches, which is then input to the BCE Classifier to retrieve the image-pair attention maps from the BCE loss. Given this, we enforce the spatial constraint objective to ensure spatial consistency of attentive regions across the two images in the input pair. 
 
We optimize our proposed CASN for all the objectives described here jointly, with the overall CASN training objective given as:
\begin{equation}
    L = L_{ide} + \lambda_1 L_{ia} + \lambda_2 L_{sa}
\label{eq:train_loss}
\end{equation}
where $L_{ide}$ is the IDE classification loss, $L_{ia}$ is the identification attention loss, and $L_{sa}$ is the Siamese attention loss. Note that the feature extraction blocks across the two branches in Figure \ref{fig:intro} share weights. The proposed CASN addresses all problems discussed previously in a principled fashion, allowing us to (a) generate attention maps with attention consistency, (b) learn identity-aware invariant representations by design, and (c) use attention maps during inference for identities not seen during training. Furthermore, compared to existing attention mechanisms employed in person re-id, our framework is flexible by design in that it can be used in conjunction with any base architecture or baseline re-id algorithm. For instance, in Section~\ref{sec:exp}, we show performance improvements with both the IDE \cite{IDE_CVPR16} and
the PCB \cite{sunPCB_ECCV18} baselines. Furthermore, we only need identity labels during training (which are used by competing algorithms as well), but crucially, do not need any specially designed architecture sub-modules to make attention a part of the learning process.

\section{Experiments and Results}
\label{sec:exp}

\noindent \textbf{Datasets.} We use Market-1501 \cite{Market1501_ICCV15}, CUHK03-NP \cite{deepReID_CVPR14, Rerank_CVPR17}, and DukeMTMC-ReID \cite{zhengDuke_ICCV17, ristaniDuke_ECCV16}. Market-1501 \cite{Market1501_ICCV15} collects person images from 6 camera views, containing 12,936 training images with 751 different identities. Gallery and query sets have 19,732 and 3,368 images respectively with 750 different identities. CUHK03-NP is a new training-testing split protocol for CUHK03 \cite{deepReID_CVPR14}, first proposed in \cite{Rerank_CVPR17}, splitting the training and testing sets into 767 and 700 identities. DukeMTMC-ReID \cite{zhengDuke_ICCV17} is an image-based re-id dataset generated from DukeMTMC \cite{ristaniDuke_ECCV16} that randomly splits training and testing sets equally into 702 identities. 

\noindent \textbf{Implementation Details.} We resize all images to 288$\times$144, use SGD with momentum of 0.9, learning rate of 0.03, and a total of 40 epochs, with the learning rate decreased by a factor of 10 at epoch 30. The parameter $\alpha$ in Equation \ref{eq:Spatial_cons} is set to 0.2, and $\lambda_1$ and $\lambda_2$ in Equation \ref{eq:train_loss} are set to 0.5 and 0.05 respectively. For the PCB baseline, we follow the same protocol as in \cite{sunPCB_ECCV18} and resize images to 384$\times$128. We set the batch size to 16, use two NVIDIA GTX-1080Ti GPUs, and implement all code in Pytorch \cite{pytorch}.

\noindent \textbf{Evaluation Protocol.} After training, we use query and gallery as pair inputs to obtain attention maps from BCE classifier predictions. The $l_2$ distance of the attention maps (Equation \ref{eq:Spatial_cons} in Section \ref{sec:SiaAtt}) and $l_2$ distance of the feature vectors are normalized and summed for final ranking. We report the rank-1 Cumulative Match Characteristic (CMC) and mean average precision (mAP) results.

\subsection{Comparison to the State of the Art}
In Tables \ref{table:stateofArt_CUHK03} and \ref{table:stateofArt_DukeMarket}, we compare the performance of our method with several recently proposed algorithms applied to the CUHK03-NP, DukeMTMC-ReID, and Market-1501 datasets. Note that all our results are evaluations without re-ranking \cite{Rerank_CVPR17} and the PCB \cite{sunPCB_ECCV18} architecture as the backend. 

\noindent \textbf{CUHK03-NP.} We report experimental results on both detected and labeled person images. The new train-test split, containing only around 7,300 training images, is much more prone to overfitting when compared to the other datasets. However, results show that our method surpasses the state of the art substantially for rank-1 (+4.7\%, +5.7\%) on detected and labeled sets respectively, demonstrating the strong generalization ability of the CASN. More crucially, compared to a recently proposed attention-based method, HA-CNN \cite{LiHACNN_CVPR18}, our CASN achieves 29.8\% and 25.8\% rank-1 and mAP improvements (on detected sets) respectively. 

\begin{table}[!h]
	\caption{CUHK03-NP (detected and labeled).}
		\vspace{-0.3cm}
	\centering
	\scalebox{0.88}{
		\setlength{\extrarowheight}{.2em}
		\begin{tabular}{|p{3cm}|cc|cc|}
		    \hline
		     & \multicolumn{2}{c|}{Detected} & \multicolumn{2}{c|}{Labeled} \\
			\cline{2-5}
			& R-1 & mAP & R-1 & mAP \\
			\hline
			BoW+XQDA \cite{Market1501_ICCV15} & 6.4\% & 6.4\% & 7.9\% & 7.3\%\\
			LOMO+XQDA \cite{LOMO_XQDA_CVPR15} & 12.8\% & 11.5\% & 14.8\% & 13.6\% \\
			\hline
			IDE \cite{Zheng_overview_CoRR16}& 21.3\% & 19.7\% & 22.2\% & 21.0\% \\
			PAN \cite{zhengPedestrian_CSVT18}& 36.3\% & 34.0\% & 36.9\% & 35.0\%\\
			DPFL \cite{DPFL_ICCVW17} & 40.7\% & 37.0\% & 43.0\% & 40.5\% \\
			HA-CNN \cite{LiHACNN_CVPR18} & 41.7\% & 38.6\% & 44.4\% & 41.0\% \\
			MLFN \cite{MLFN_CVPR18} & 52.8\% & 47.8\% & 54.7\% & 49.2\% \\
			DaRe+RE \cite{DaRe_CVPR18} & 63.3\% & 59.0\% &  66.1\% & 61.6\% \\
			PCB+RPP \cite{sunPCB_ECCV18} & 63.7\% & 57.5\% & - & -\\
 			MGN \cite{MGN_MM18}  & 66.8\% & 66.0\% & 68.0\% & 67.4\%\\
			\hline
			CASN (IDE) & 57.4\% & 50.7\% & 58.9\% & 52.2\% \\ %
			\textbf{CASN (PCB)} & \textbf{71.5\%} & \textbf{64.4}\% & \textbf{73.7\%} & \textbf{68.0\%} \\ %
			\hline
	\end{tabular}}
\label{table:stateofArt_CUHK03}
\end{table}

\begin{table}[!h]
	\caption{DukeMTMC-ReID and Market-1501 (SQ).}
	\vspace{-0.3cm}
	\centering
	\scalebox{0.85}{
		\setlength{\extrarowheight}{.2em}
		\begin{tabular}{|p{3cm}|cc|cc|}
		    \hline
		     & \multicolumn{2}{c|}{DukeMTMC-ReID} & \multicolumn{2}{c|}{Market-1501} \\
			\cline{2-5}
			& R-1 & mAP & R-1 & mAP \\
			\hline
			BoW+KISSME \cite{Market1501_ICCV15} & 25.1\% & 12.2\% & 44.4\% & 20.8\%\\
			LOMO+XQDA \cite{LOMO_XQDA_CVPR15} & 30.8\% & 17.0\% & 43.8\% & 22.2\% \\
			\hline
			SVDNet \cite{SVDnet_ICCV17} & 76.7\% & 56.8\%  &82.3\% &62.1\%\\
			HA-CNN \cite{LiHACNN_CVPR18} & 80.5\% & 63.8\% & 91.2\%& 75.7\%\\
			DuATM \cite{DuATM_CVPR18} & 81.8\% & 64.6\% & 91.4\%&76.6\% \\
			PCB+RPP \cite{sunPCB_ECCV18} & 83.3\% & 69.2\% & 93.8\% &81.6\% \\
			DNN\_CRF \cite{DNN_CRF_CVPR18} & 84.9\% & 69.5\% &- &-\\
 			\textbf{MGN} \cite{MGN_MM18}  & \textbf{88.7\%} & \textbf{78.4\%} & \textbf{95.7\%} & \textbf{86.9\%}\\
			\hline
			CASN (IDE) & 84.5\% & 67.0\% & 92.0\% & 78.0\% \\ %
			CASN (PCB) & 87.7\% & 73.7\% & 94.4\% & 82.8\% \\ %
			\hline
	\end{tabular}}
\label{table:stateofArt_DukeMarket}
\end{table}

\begin{figure*}[h!]
	\centering
	\subfloat[Attention maps retrieved from BCE loss (training)]{\includegraphics[width=0.9\linewidth, height=1.2cm]{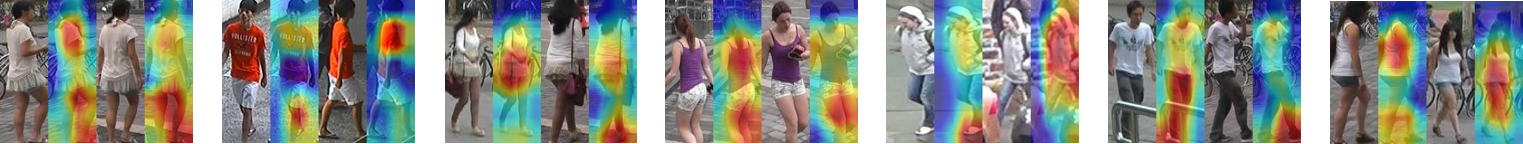}}\\
	\subfloat[Attention maps retrieved from BCE loss with Siamese Attention loss (training)]{\includegraphics[width=0.9\linewidth, height=1.2cm]{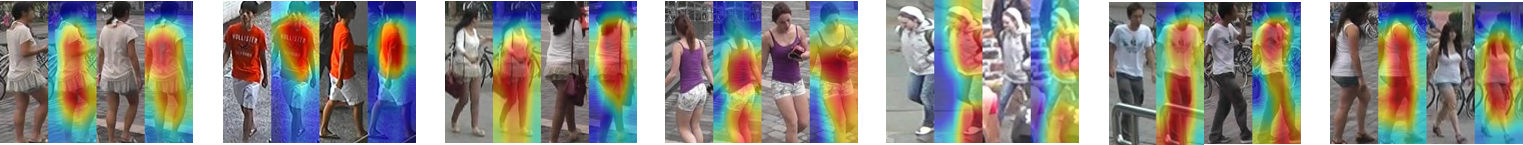}}\\
	\subfloat[Attention maps retrieved from model trained with Siamese Attention (Rank 1 gallery match for query images)]{\includegraphics[width=0.9\linewidth, height=1.2cm]{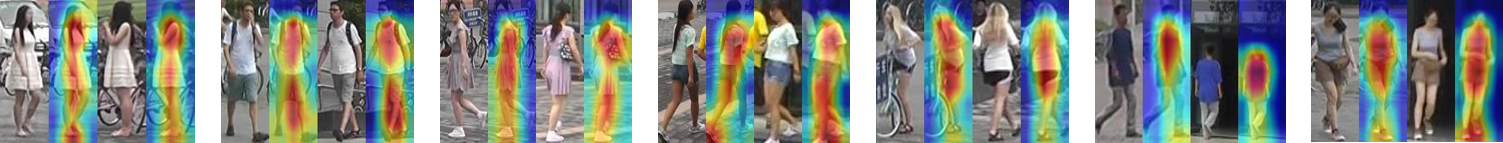}}\\
		\vspace*{-0.2cm}
	\caption{Demonstrating the efficacy of the proposed Siamese attention by means of attention maps for same-person images.} 
	\label{fig:AttMap_abl}
\end{figure*}

\noindent \textbf{DukeMTMC-ReID.} We report competitive results in Table \ref{table:stateofArt_DukeMarket}. Again, compared to recently proposed attention-based methods, HA-CNN \cite{LiHACNN_CVPR18} and DuATM \cite{DuATM_CVPR18}, our CASN achieves 7.2\% and 5.9\% rank-1 accuracy improvements and 9.9\% and 9.1\% mAP improvements respectively. 

\noindent \textbf{Market-1501.} We report competitive results with CASN in Table \ref{table:stateofArt_DukeMarket}. However, compared to recently proposed attention-based methods, e.g., HA-CNN \cite{LiHACNN_CVPR18} and DuATM \cite{DuATM_CVPR18} (shown in the table), and CAN \cite{CAN_TIP17} (R-1: 60.3\%, mAP: 35.9\%), HPN \cite{liuHPnet_ICCV17} (R-1: 76.9\%), MSCAN \cite{LiMSCAN_CVPR17} (R-1: 80.3\%, mAP: 57.5\%) our method produces much higher results with both rank-1 and mAP. 

As can be noted from these results, the proposed CASN substantially outperforms existing attention-based methods for re-id. More importantly, unlike these competing attention-based methods, CASN does not require any specially designed deep architecture for modeling attention, relying only on identity labels for supervision. This allows the CASN to be highly flexible for use in conjunction with any baseline CNN architecture, such as VGGNet \cite{VGGnet_CoRR14}, DenseNet \cite{huangDense_ICCV17}, or SqueezeNet \cite{SqueezeNet_16}. For instance, with DenseNet and the IDE baseline, CASN achieves a rank-1 and mAP performance of 57.2\% and 52.0\% respectively on CUHK03-NP (detected), which are close to CASN's results with ResNet50 and IDE, discussed next.

\subsection{Ablation Study and Discussion}
In this section, we further study the role of the identification attention and Siamese attention mechanisms individually, and how they influence the performance of the CASN. In Table \ref{table:ablation}, we report evaluation results of our proposed model on CUHK03-NP (detected), DukeMTMC-ReID and Market-1501, starting from baseline IDE and PCB architectures and working up to the full CASN model. From Table \ref{table:ablation}, we can see clear performance improvements over the baseline with individual attention modules. For instance on CUHK03-NP, IDE+IA improves the rank-1 and mAP performance of baseline IDE by 9.0\% and 9.2\% whereas IDE+SA improves the rank-1 accuracy by 9.4\% and 10.2\% respectively. This provides evidence for our initial hypothesis that spatial localization, via end-to-end trainable attention mechanisms, should be an important and integral part of the framework design. Furthermore, adding both attention modules improves performance as measured by both rank-1 accuracy and mAP, demonstrating the importance of using both identification and Siamese modules.

\begin{table}[!h]
	\caption{Ablation study. IA: Identification Attention, SA: Siamese Attention, SQ: Single-Query.}
	\centering
	\scalebox{0.73}{
	\setlength{\extrarowheight}{0.2em}
	\begin{tabular}{|p{1.7cm}|cc|cc|cc|}
		\hline
        Loss type & \multicolumn{2}{c|}{CUHK03-NP} & \multicolumn{2}{c|}{DukeMTMC-ReID} & \multicolumn{2}{c|}{Market-1501 (SQ)} \\
        \cline{2-7}
		& R-1 & mAP & R-1 & mAP & R-1 & mAP \\
		\hline
		IDE \cite{sunPCB_ECCV18} & 43.8\% & 38.9\% & 73.2\% & 52.8\% & 85.3\% & 68.5\%  \\
		IDE + IA & 54.8\% & 48.1\% & 83.2\% & 66.0\% & 91.0\% & 76.9\% \\
		IDE + SA & 55.2\% & 49.1\% & 83.5\% & 66.0\% & 91.6\% & 77.7\% \\
        \hline
		\textbf{CASN(IDE)} & \textbf{57.4\%} & \textbf{50.7\%} & \textbf{84.5\%} & \textbf{67.0\%} & \textbf{92.0\%} & \textbf{78.0\%} \\
		\hline
		PCB \cite{sunPCB_ECCV18} & 61.3\% & 54.2\% & 81.7\% & 66.1\% & 92.4\% & 77.3\%  \\
		PCB + IA & 68.5\% & 62.4\% & 87.3\% & 73.4\% & 93.9\% & 81.8\% \\
		PCB + SA & 69.9\% & 64.2\% & 86.8\% & 73.5\% & 94.1\% & 82.6\% \\
		\hline
	    \textbf{CASN(PCB)} & \textbf{71.5\%} & \textbf{64.4\%} & \textbf{87.7\%} & \textbf{73.7\%} & \textbf{94.4\%} & \textbf{82.8\%} \\
		\hline
\end{tabular}}
	\label{table:ablation}
\end{table}

\begin{figure}[hbtp!]
	\centering
	\vspace{-0.25cm}
	\includegraphics[width=0.8\linewidth]{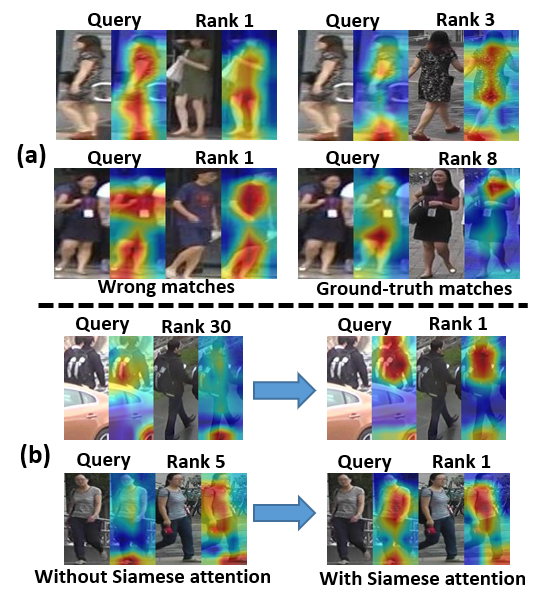}%
	\caption{(a) Our attention maps can explain wrong (high rank, e.g., rank 1) and ground-truth matches (low rank, e.g., rank 3). (b) Siamese attention gives rank improvements, providing reasoning with attention consistency. } 
	\label{fig:insights}
	\vspace{-0.5cm}
\end{figure}

Comparisons of the attention maps acquired from the models trained with BCE loss and BCE loss with Siamese Attention loss are shown in Figure \ref{fig:AttMap_abl}(a-b). Clearly, with the proposed Siamese attention mechanism, we obtain more consistent attention maps of the same person image pair in Figure \ref{fig:AttMap_abl}(b) compared to Figure \ref{fig:AttMap_abl}(a). Furthermore, we also demonstrate these attention maps for the testing image pairs in Figure \ref{fig:AttMap_abl}(c), where we again see attention consistency among the query and retrieved gallery images. These examples demonstrate the effectiveness of our proposed Siamese attention mechanism, and also provide a powerful interpretability tool. With such attention maps, we can now explain why our Siamese network predicts a certain input image pair to be similar or dissimilar, leading to intuitive explanations for person re-id. In more detail, Figure \ref{fig:insights}(a) shows two query images (one on each row), along with their rank-1 (left column) and ground-truth matches (right column). Each rank-1 match is a wrong match (failure case) while the ground-truth has a lower rank, and we can understand the reasoning from our attention maps. For instance, on the first row, we see reasonable attention consistency between the query and rank-1 (notice both show women in dresses), explaining why the wrong match was ranked 1, unlike the ground-truth, where we see attention focused on different regions, leading to lower rank (rank 3 in this example). In Fig~\ref{fig:insights}(b), we demonstrate the efficacy of our proposed Siamese attention (two examples, one on each row). The left column shows \{query, ground-truth\} and the ground truth's rank without Siamese attention. The right column shows these results with Siamese attention. We can see that Siamese attention results in better attention consistency, which is also reflected in the improved rank.

\section{Conclusions}
We proposed the first learning architecture that integrates attention consistency modeling and Siamese representation learning in a joint learning framework, called the Consistent Attentive Siamese Network (CASN), for person re-id. Our framework provides for principled supervisory signals that guide our model towards discovering consistent attentive regions for same-identity images while also learning identity-aware invariant representations for cross-view matching. We conducted extensive evaluations on three popular person re-id datasets and demonstrated competitive performance. While computing attention as in Section~\ref{sec:SiaAtt} is specific to standing poses that are common in existing benchmarks, our framework is extensible to enforce different kinds of consistency given data- or domain-specific priors for real-world generalizability.  

\paragraph{Acknowledgements} This material is based upon work supported by the U.S. Department of Homeland Security under Award Number 2013-ST-061-ED0001. The views and conclusions contained in this document are those of the authors and should not be interpreted as necessarily representing the official policies, either expressed or implied, of the U.S. Department of Homeland Security.



{\small
\bibliographystyle{ieee_fullname}
\bibliography{egbib}
}

\end{document}


\title{Supplemental Material: Re-Identification with Consistent Attentive Siamese Networks}

\maketitle
\begin{multicols}{2}
\section{More Experimental Results}
Figure \ref{fig:AttMap} provides more qualitative results showing consistent attentive regions for query images and their rank-1 match from gallery. The model is trained with IDE baseline with Siamese attention, corresponding to Figure 8(c) in the main paper.
\end{multicols}

\begin{figure*}[ht]
	\centering
	\includegraphics[width=0.9\linewidth, height=12cm]{figure//suppl_consist_v.png}
	\caption{Demonstration of the consistency of the attention maps for query images and their rank-1 gallery match.}
	\label{fig:AttMap}
\end{figure*}